\documentclass{article}
\usepackage{graphicx}
\usepackage{subcaption} 
\usepackage{amsmath}
\usepackage{authblk}              
\usepackage{abstract}             
\usepackage{hyperref}             
\usepackage{amsmath, amssymb}     
\usepackage{lipsum}   
\usepackage{authblk}
\usepackage{float}
\usepackage{changepage}
\usepackage{multirow}
\usepackage[numbers,sort&compress]{natbib}

\title{A Global-Local Cross-Attention Network for Ultra-high Resolution Remote Sensing Image Semantic Segmentation}

\author[1]{Chen Yi}
\author[2]{Shan LianLei}

\affil[1]{Xi'an University of Science and Technology}
\affil[2]{University of Chinese Academy of Sciences} 

\begin{document}
\maketitle
\begin{abstract}
With the rapid development of ultra-high resolution (UHR) remote sensing technology, the demand for accurate and efficient semantic segmentation has increased significantly. However, existing methods face challenges in computational efficiency and multi-scale feature fusion. To address these issues, we propose GLCANet (Global-Local Cross-Attention Network), a lightweight segmentation framework designed for UHR remote sensing imagery.GLCANet employs a dual-stream architecture to efficiently fuse global semantics and local details while minimizing GPU usage. A self-attention mechanism enhances long-range dependencies, refines global features, and preserves local details for better semantic consistency.
A masked cross-attention mechanism also adaptively fuses global-local features, selectively enhancing fine-grained details while exploiting global context to improve segmentation accuracy. Experimental results show that GLCANet outperforms state-of-the-art methods regarding accuracy and computational efficiency. The model effectively processes large, high-resolution images with a small memory footprint, providing a promising solution for real-world remote sensing applications.
\end{abstract}

\newpage
\section{Introduction}

With the rapid advancement of high-resolution Earth observation technology, semantic segmentation of ultra-high-resolution (UHR) remote sensing imagery has become a core technique for land cover classification and urban change detection\cite{bonafilia2020sen1floods11}. However, the multi-scale object distribution (e.g., the coexistence of small roads and large building complexes) and complex geometric structures (e.g., irregular farmland boundaries) in UHR imagery pose dual challenges to existing segmentation models. On the one hand, single-stream architectures struggle to model global contextual information and local fine-grained details jointly. On the other hand, dual-stream networks suffer from feature coupling issues, leading to parameter redundancy and computational inefficiency\cite{wang2020deep}.

Current studies mainly address these challenges through multi-scale feature fusion. A representative approach, DeepLabv3+\cite{chen2018encoder}, employs Atrous Spatial Pyramid Pooling (ASPP) to capture multi-scale contextual information. However, its rigid pyramid structure fails to adapt to the scale variations in remote sensing imagery. Zhu et al.\cite{zhang2018road} introduced a cascade attention mechanism to enhance road extraction, but it does not resolve the memory bottleneck associated with high-resolution processing. Zhang et al.\cite{zhang2018context} adopted a dual-branch architecture to separate global and local feature extraction, but the simple feature concatenation results in semantic inconsistencies. These methods still face significant limitations in balancing segmentation accuracy and computational efficiency.

To address the issues above, this paper proposes an innovative deep learning model, GLCANet (Global-Local Cross-Attention Network), which integrates a dual-stream collaborative architecture with a cross-modal attention mechanism. Explicitly designed for UHR remote sensing image segmentation, GLCANet effectively combines global semantic information with local fine-grained details through parallel dual-stream feature extraction and an efficient feature fusion mechanism. The international branch employs strategic downsampling to capture macro-level contextual information, while the local branch focuses on high-resolution image patches to extract fine-grained texture features. To further enhance multi-scale feature fusion, this paper introduces the Global-Local Cross-Attention Fusion Module (GLCA-FM), which integrates a self-attention mechanism to model long-range dependencies and a masked attention mechanism to preserve spatial structure integrity. 

Experimental results demonstrate that GLCANet significantly outperforms traditional methods regarding segmentation accuracy while maintaining superior computational efficiency and memory utilization. The model effectively processes UHR remote sensing imagery with reduced resource consumption, making it a promising solution for large-scale, high-resolution segmentation tasks.

It is remarkable to witness the advancements in semantic segmentation of remote sensing imagery driven by Convolutional Neural Networks (CNNs). This cutting-edge technology has enabled the precise labeling of pixels based on ground information of interest, paving the way for many practical applications such as environmental assessment, crop monitoring, natural resource management, and digital mapping. With the progression of photography and sensor technology, we now have access to high-resolution remote-sensing images that facilitate the possibility of automatic semantic segmentation.

Our primary contributions are encapsulated as follows:
\begin{itemize}
\item Proposed a lightweight semantic segmentation framework, GLCANet: To address the computational resource constraints in UHR remote sensing image processing, a dual-stream collaborative architecture is designed. This architecture enables efficient fusion of global semantic information and local fine-grained features while maintaining strict GPU memory constraints, thereby improving both segmentation accuracy and computational efficiency.
\item Introduced a self-attention mechanism to optimize feature representation: After the global and local branches extract their respective features, a self-attention mechanism is incorporated to enhance long-range dependency modeling. This mechanism refines low-resolution global features by enriching contextual information and improves high-resolution local features by enhancing their expressiveness. As a result, it ensures better semantic consistency and boundary detail preservation, providing a more robust foundation for subsequent feature fusion.
\item Designed a masked cross-attention mechanism to enhance global-local feature interaction: A masked cross-attention mechanism is proposed to achieve adaptive fusion of global and local features. The model effectively extracts key fine-grained details while leveraging global contextual guidance to refine local feature representations by introducing a masking matrix to constrain attention computation. This approach significantly enhances multi-scale feature fusion efficiency, thereby improving segmentation accuracy.
\end{itemize}

\section{RELATED WORK}

Semantic segmentation, empowered by Convolutional Neural Networks (CNNs), has made significant strides across a spectrum of applications, including saliency detection \cite{wang2018detect}, medical imaging segmentation \cite{ronneberger2015u}, and road scene understanding. The architectural paradigm typically features an encoder-decoder framework, where the encoder condenses local image information, and the decoder restores spatial details \cite{badrinarayanan2017segnet}. Innovations such as deformable convolution \cite{chen2014semantic}, dilated convolution \cite{chen2018encoder}, and pyramid pooling modules have been integrated to enhance local feature extraction and contextual modeling at various scales. State-of-the-art CNN architectures like HRNet \cite{wang2020deep} and RefineNet \cite{lin2017refinenet} have further propelled feature extraction capabilities. And others also do some meanningful works in segmentation \cite{densenet,uhrsnet,tgrs1,decouple,mbnet,tgrs2,liminglong,zhaoyuzhong,acmmm,boosting,data,zhaoguiqin,lifelong,cognitive,dlnet,rs2,fusing,ldnet,organizing,edge,binary,energy,boosting,synthetic,dynrsl,flexdataset,gmm,llmcot,geogrambench,geolocsft,f2net}.

Fully Convolutional Networks (FCNs) have laid the groundwork for deep learning in semantic segmentation. Pioneering works like U-Net \cite{liu2017image, liu2020connecting} utilize skip connections within an encoder-decoder structure to harmonize low-level and high-level features. Other notable models include DeconvNet \cite{noh2015learning}, which shares structural similarities with SegNet \cite{badrinarayanan2017segnet}, and RefineNet, which incorporates a multipath optimization module for recursive multiscale feature exploitation. The Feature Pyramid Network (FPN) employs a top-down fusion approach to amalgamate multiscale semantic insights. DeepLab \cite{chen2014semantic, chen2017deeplab, yu2015multi} stands out for its use of dilation convolution to broaden the filter's field of view. Despite these advancements, a focus on local features can limit the modeling of global structural and semantic information, and the memory-intensive nature of these methods poses challenges, particularly with ultra-high-resolution images.

The pursuit of real-time semantic segmentation has seen models like ENet \cite{paszke2016enet} adopt asymmetric encoder-decoder architectures with early downsampling to curtail computational expenses. ICNet \cite{zhao2018icnet} and BiSeNet amalgamates multi-resolution feature maps through cascades and parallel branches to address labeling and compression challenges. However, these approaches may falter when confronted with ultra-high-resolution images' memory and computational demands, potentially leading to diminished performance or prolonged execution times. Moreover, they grapple with category imbalance, where small target classes may be underserved, affecting the model's overall efficacy.

In the realm of remote sensing image semantic segmentation, the task involves meticulous classification of land cover and use categories, such as roads \cite{ding2016road}, buildings \cite{ding2021adversarial}, and water bodies \cite{duan2019multiscale}. Spatial precision is paramount for remote sensing applications, underscoring the critical need for accurate segmentation.

While effective, GLNet's approach to constructing global and local branches through the fusion of downsampled and cropped image blocks may not fully discriminate the significance of each image block, leading to excessive feature integration. To counter this, we propose the Adaptive Weighting Multi-Field-of-View Convolutional Neural Network (AWMF-CNN), which has three parallel branches that process multi-scale image blocks from target regions. AWMF-CNN adeptly addresses classification inaccuracies through weighted fusion, enriching the spatial context with multi-resolution inputs. However, integrating substantial external networks in AWMF-CNN escalates the parameter count exponentially, impacting operational efficiency.

Attention mechanisms have become prevalent in the semantic segmentation of remote-sensing images. Innovations such as the Squeeze-and-Excitation (SE) block, as extended in \cite{ding2020lanet}, focus on semantic alignment in the spatial dimension, bridging the gap between high-level and low-level features. However, the SE approach may not fully account for spatial layouts and contextual relationships within remote sensing imagery. High-resolution images further amplify computational loads, potentially restricting the practical application efficiency of networks like HRNet. Furthermore, these methods often necessitate extensive labeled datasets to overcome the diversity and complexity inherent in remote sensing images. However, procuring large-scale, high-quality annotated data for remote sensing can be particularly challenging, especially for specific regions or tasks.

\begin{figure}[H]
\centering
\includegraphics[width=15.5cm]{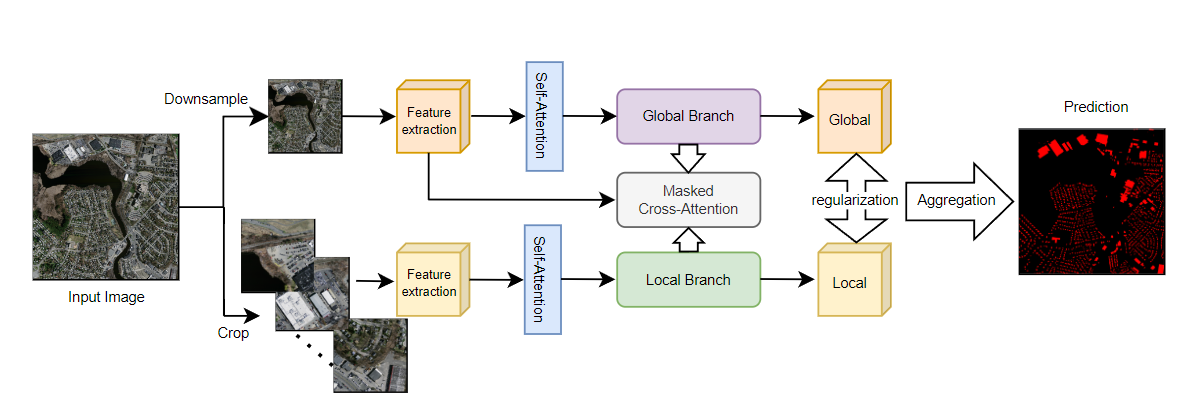}
\caption{Overview of the Model. The global and local branches leverage downsampled and cropped images, respectively. The deep feature maps are enhanced with a Cross-Attention mechanism to bolster our global-local collaboration. The final segmentation is derived from the high-level feature maps of the two branches. \label{fig:1}}
\end{figure}

\subsection{Lightweight Remote Sensing Image Segmentation Network}
Semantic segmentation of ultra-high-resolution (UHR) remote sensing imagery presents challenges due to high computational complexity and data volume. Traditional deep learning models require dense computations on full-resolution images, resulting in excessive GPU memory consumption and low inference efficiency, limiting their practicality for UHR image processing.

Researchers have developed lightweight segmentation frameworks to reduce computational costs and accelerate inference to address this. ICNet\cite{zhao2018icnet} adopts a cascade inference strategy, lowering input resolution to balance accuracy and efficiency, but suffers from detail loss in boundaries and small-object recognition. ENet\cite{paszke2016enet} leverages network pruning and dimensionality reduction to improve efficiency, but at the cost of feature extraction, leading to lower segmentation accuracy, especially in multi-scale scenarios where maintaining semantic consistency is crucial.

Despite these advancements, existing methods struggle to effectively integrate global and local features. UHR remote sensing imagery exhibits multi-scale characteristics:

\begin{itemize}
\item Large objects (e.g., lakes, forests) require global information for category consistency.
\item Small objects (e.g., roads, bridges, buildings) rely on high-resolution local features for structural detail.
\end{itemize}
Traditional lightweight approaches depend heavily on downsampling for global features, which sacrifices fine details, especially in complex textures and boundary regions. Conversely, relying only on local high-resolution features may lead to semantic inconsistency, affecting segmentation performance. Thus, efficiently fusing global and local information under constrained computational resources remains a key challenge in UHR segmentation.

BiSeNet\cite{yu2018bisenet} integrates a Fast Spatial Path and Context Path to balance efficiency and contextual modeling. However, it prioritizes feature extraction speed over global-local interaction, limiting its effectiveness for precise UHR segmentation.

In recent years, image segmentation has become a crucial task in computer vision, encompassing several subtasks such as semantic segmentation, instance segmentation, and panorama segmentation. Although convolutional neural network methods like FCN, DeepLab, and Mask R-CNN have achieved remarkable success in image segmentation, they still face limitations in establishing dependencies between distant pixels. In contrast, the Transformer network possesses a robust global information interaction capability, enabling the rapid establishment of a global sensory field and improving scene understanding accuracy. With the Transformer network, the segmentation task is better equipped to capture long-range relationships between pixels, overcoming the shortcomings of traditional methods in dealing with long-range pixel dependencies. The positions Transformer has significant potential in image segmentation, leading to more accurate segmentation results and enhanced scene understanding.

Recently, the application of Transformers to computer vision tasks has garnered extensive research interest. In \cite{dosovitskiy2020image}, the Vision Transformer (ViT) is introduced for image classification, demonstrating that a pure Transformer can be an alternative to CNNs for image recognition tasks. The corresponding attention modules are gradually being applied to semantic segmentation networks. In \cite{wang2021max}, a two-channel Transformer is proposed for panorama segmentation, which includes a pixel channel for segmentation and a memory channel for category prediction. The Transformer is utilized to communicate information between the two channels. The RAFCN structure proposed in \cite{wang2018non} processes features in both the channel dimension and spatial dimension, unlike DANet, which performs attentional reinforcement and deconvolution for three different levels of abstract information, respectively. In \cite{bazi2021vision}, the Visual Transformer demonstrates an advantage over CNNs in scene categorization of RSIs. In \cite{chen2021remote}, a diachronic Transformer is introduced for change detection in RSIs. Bitemporal semantic features are disambiguated and concatenated, subsequently enriched with global semantic correlations via Transformer. Although the Transformer shows some advantages in processing remote sensing images, there are some challenges: the Transformer model usually requires more computational resources and time for training, leading to higher computational costs and training times than traditional CNN models. The information in some remote sensing images may be more localized, and Transformers may not perform as well as some specially designed CNN architectures in dealing with local features.

\subsection{Application of Self-Attention Mechanism in Remote Sensing Image Segmentation}
In semantic segmentation tasks, traditional convolutional neural networks (CNNs) rely on local receptive fields for feature extraction, making capturing long-range dependencies between distant pixels difficult. However, in ultra-high-resolution (UHR) remote sensing imagery, objects exhibit complex structures and significant scale variations, where long-range spatial dependencies often exist across different regions. For instance, road networks span entire images; urban buildings follow specific spatial layouts and agricultural fields exhibit regular patterns. If segmentation relies solely on CNN-based feature extraction, it may lead to inconsistent segmentation across large-scale objects, loss of fine details, and inaccurate object boundaries. Therefore, enhancing global feature representation and improving long-range dependency modeling is a key challenge in remote sensing image segmentation.

In recent years, self-attention mechanisms have been widely adopted in computer vision tasks, particularly semantic segmentation. By capturing long-range dependencies, self-attention overcomes the local receptive field limitations of CNNs, significantly improving global feature utilization. Among representative approaches, Non-Local Networks\cite{wang2018non} employ non-local operations to compute similarity relationships between all pixels in a feature map, enabling global feature interaction. This allows distant spatial regions to influence each other directly, enhancing object consistency in segmentation.

Building on this foundation, Transformer architectures enhance the use of self-attention mechanisms in semantic segmentation. The Swin Transformer\cite{liu2021swin}, for instance, introduces a sliding window self-attention mechanism that balances computational efficiency with improved feature modeling for high-resolution images. In contrast to CNNs that employ fixed receptive fields, the Swin Transformer progressively facilitates global feature interaction through a hierarchical approach that combines local window computations with window shifting. This enables more accurate segmentation of ultra-high-resolution images. Furthermore, SegFormer\cite{xie2021segformer} merges self-attention mechanisms with a lightweight decoder, optimizing computational efficiency and enabling high-precision segmentation at reduced computational costs in remote sensing image analysis.

Despite these advancements in global feature modeling, existing methods still face challenges. Most approaches do not explicitly address the imbalance between global and local feature representations, making it challenging to capture fine details in the low-resolution global feature branch while effectively integrating global context in the high-resolution local feature branch. In other words, global features may lack fine-grained details, while local features may suffer from missing global semantic guidance, ultimately impacting segmentation accuracy.

For instance, in remote sensing imagery, segmentation of large-scale objects (e.g., forests, lakes) requires global consistency constraints. In contrast, precise boundary delineation of roads or buildings relies on high-resolution local features. Segmentation models may struggle with category inconsistencies and blurred boundaries if global and local features are not effectively fused. Addressing these limitations remains a crucial challenge for improving semantic segmentation performance in ultra-high-resolution remote sensing imagery.

\subsection{Application of Cross-Attention Mechanism in Feature Fusion}
Feature fusion is critical in multi-scale semantic segmentation tasks, aiming to establish effective information interaction between scales to enhance target recognition and localization. Traditional feature fusion methods primarily include Feature Pyramid Networks (FPN)\cite{lin2017feature} and Atrous Spatial Pyramid Pooling (ASPP)\cite{chen2017deeplab}. FPN employs a top-down multi-scale fusion strategy, gradually enhancing the semantic representation of lower-level features and significantly improving small-object detection. However, its fusion method mainly relies on simple hierarchical weighting or feature concatenation, which lacks adaptability in adjusting feature importance. On the other hand, ASPP utilizes atrous convolutions with different sampling rates to facilitate multi-scale feature interaction and enhance feature integration. Nevertheless, it still depends on fixed convolution operations, making it challenging to adjust dependencies based on task-specific requirements dynamically.

In recent years, the cross-attention mechanism has been widely adopted in semantic segmentation to achieve more adaptive feature fusion between different feature streams. Unlike traditional weighted averaging or simple concatenation, cross-attention dynamically assigns weights based on the relevance of global and local information, thereby improving the quality of fused features. For example, CoTNet\cite{zhao2024cooperative} integrates convolution and attention mechanisms, learning feature relationships within a local receptive field while incorporating non-local attention mechanisms to enhance feature interaction. Similarly, CrossViT\cite{chen2021crossvit} employs cross-attention within multi-scale Transformer branches, enabling low-resolution global features to effectively guide high-resolution local feature representation, thus enhancing segmentation accuracy.

Although these methods have improved feature fusion capabilities, they face challenges in ultra-high-resolution (UHR) remote sensing image segmentation. First, the high computational cost of UHR images significantly increases the computational burden of Transformer-based models, impacting inference efficiency. Second, due to the large-scale variations of land cover objects in remote sensing images, naïve cross-attention mechanisms struggle with spatial alignment between global and local information. This misalignment can result in blurred segmentation boundaries and inconsistent category predictions.

Therefore, optimizing the cross-attention mechanism for UHR image segmentation remains an important research direction. This focus is on improving computational efficiency while enhancing the interaction between global and local features to ensure more precise and robust segmentation.

\section{METHODOLOGY}

\subsection{Architecture of GLCANet}

This paper proposes a lightweight dual-branch collaborative network, GLCANet, designed to achieve high-precision semantic segmentation of ultra-high-resolution (UHR) remote-sensing images under limited GPU resources. As illustrated in Figure 1, GLCANet consists of three key components: the Global Branch, the Local Branch, and the Global-Local Cross-Attention Fusion Module (GLCA-FM). Specifically, the Global Branch rapidly captures macro-scale contextual information through a downsampling strategy and employs a modified ResNet-50 as the backbone to extract multi-scale semantic features. In contrast, the Local Branch retains the original or high-resolution input by processing the image in local patches, focusing on extracting fine-grained details such as edges and textures, thereby preserving spatial resolution. To bridge the semantic and detail gaps between the two branches, GLCA-FM introduces a cross-attention mechanism that effectively facilitates cross-scale information interaction and adaptive feature fusion, ensuring high segmentation accuracy and consistency while maintaining computational efficiency.

\subsection{Global Branch}
The Global Branch is designed to efficiently extract macro-scale semantic contextual information at a lower resolution, thereby reducing GPU memory consumption and enhancing the modeling of long-range dependencies in ultra-large remote sensing images. Specifically, an ultra-high-resolution input image is first downsampled to produce a low-resolution version $I^{lr}\in \mathbb{R}^{{h_1 \times w_1}}$, where $h_1$,$w_1\ll H,W$. The downsampled image is then processed through a series of cascaded convolutional blocks—employing the same backbone architecture as the Local Branch—to progressively extract multi-scale contextual features while preserving the global semantic layout of the scene.
By combining resolution reduction with hierarchical feature extraction, the Global Branch significantly lowers GPU memory requirements while maintaining the capacity to understand large-scale scene structures and the spatial distribution of various land cover categories. Moreover, during training and inference, the Global Branch operates on the complete downsampled image rather than random crops, ensuring that the generated global prior information remains continuous and consistent. Such global contextual information is a strong basis for effective cross-attention integration with the Local Branch.

\subsection{Local Branch}

The Local Branch processes high-resolution images by dividing them into smaller patches and extracting features from each patch. It is designed explicitly for fine-grained feature extraction in high-resolution local regions, enabling the capture of detailed textures, edge structures, and small objects. Unlike the Global Branch, which applies downsampling, the Local Branch directly processes image patches at nearly the original resolution, preserving intricate local details. These high-resolution patches are obtained by cropping local regions from large-scale remote sensing images, thereby preventing the loss of fine-scale information due to downsampling.  

A Local Enhancement Module is incorporated within the Local Branch to refine and amplify fine-grained details. This module typically consists of shallow but high-resolution convolutional networks, potentially including multiple convolutional layers or residual blocks. These layers avoid significant downsampling to maintain the spatial integrity of the extracted features. On the one hand, the Local Enhancement Module utilizes convolutional filters to extract subtle features such as textures and edges, ensuring that small objects' shapes and boundaries are preserved and highlighted. On the other hand, it employs feature normalization and activation functions to suppress noise and redundant information that may be present in high-resolution inputs.  

After processing through this module, the Local Branch 
 generates a high-resolution local feature map rich in fine-grained details, including well-defined edge contours and local texture patterns. This high-resolution representation is particularly crucial for detecting small-scale or visually complex objects. For example, in remote sensing imagery, elements such as minor buildings, road details, or farmland boundaries require precise texture and geometric descriptions, which the Local Branch provides. By maintaining high-resolution processing, the Local Branch ensures that essential details are not lost due to the downsampling process applied in the Global Branch, thereby enhancing the segmentation accuracy for fine-scale structures.

\subsection{Global-Local Cross-Attention Fusion Module (GLCA-FM)}

The Global-Local Cross-Attention Fusion Module (GLCA-FM) employs a masked cross-attention mechanism to integrate global and local feature representations effectively. Within this module, one feature branch (e.g., the global feature map) serves as the key (K) and value (V). In contrast, the other branch (e.g., the local feature map) acts as the query (Q), establishing an interactive exchange between the two feature representations. Initially, global and local features undergo projection transformations into a shared query-key-value (QKV) space. Letting \(F_g\) denote the global feature map and \(F_l\) the local feature map, the projection is formulated as:

\[
Q = F_l W^Q, \quad K = F_g W^K, \quad V = F_g W^V
\]

where \(W^Q, W^K, W^V\) are learnable transformation matrices. In this formulation, the global feature map \(F_g\) is used as both key and value, while the local feature map \(F_l\) acts as the query. This enables local features to extract contextual semantic information from the global representation dynamically. Conversely, the roles can also be reversed, allowing the global feature map to query the local feature map to incorporate fine-grained spatial details, thereby enhancing the global-local feature interaction.

To compute cross-attention, the first step involves measuring the similarity between the query \(Q\) and key \(K\) using a scaled dot-product, followed by softmax normalization to obtain the attention weights:

\[
A = \text{Softmax} \left( \frac{Q K^T}{\sqrt{d_k}} \right)
\]

Where \(d_k\) is the feature dimensionality scaling factor, stabilizing gradient updates and preventing numerical instability. The attention weights are then applied to the value \(V\) to generate the fused feature representation:

\[
F_{\text{GLCA}} = A V
\]

This process ensures that local feature representations are dynamically refined based on global semantic context, enhancing semantic consistency across different regions. Simultaneously, through bidirectional cross-attention, the global feature map can selectively retrieve spatially significant fine-grained details from the local feature map, mitigating the loss of intricate spatial information that may arise when relying solely on global representations.

By leveraging global-local cross-attention, GLCA-FM significantly improves multi-scale feature representation learning. Accurate object recognition in semantic segmentation tasks necessitates global contextual understanding (e.g., land cover categories and spatial relationships) and precise boundary delineation. The integration of GLCA-FM allows the global feature map to provide holistic semantic guidance while the local feature map supplements fine details, thereby improving classification consistency and reducing misclassification errors. The fusion mechanism in GLCA-FM effectively enhances the segmentation accuracy by ensuring a cohesive integration of global information and local details, thereby overcoming the fragmentation issues inherent in traditional feature fusion approaches.

Figure 2 illustrates the details of the Self-Attention mechanism, which takes the feature maps of the global and local branches as input and produces context-aware attention maps. Figure 3 illustrates the details of the Masked Cross-Attention mechanism, which enables the network to learn the relationships between different regions in the input feature maps. With this structure, the proposed model can handle different scales and complex scenes in an image more effectively, resulting in more accurate results in segmentation tasks.

After downsampling, remote sensing images retain large-scale semantic information; however, fine-grained details such as building boundaries, road textures, and terrain contours may become blurred or lost due to reduced resolution. Traditional convolutional neural networks (CNNs) are constrained by their local receptive fields, making it difficult to recover these missing details. In contrast, the self-attention mechanism facilitates pixel-wise information interaction, allowing the model to restore key fine-grained features within the low-resolution feature maps. This enables the global feature extraction process to achieve a more precise and expressive semantic representation. Remote sensing images often contain objects with long-range spatial dependencies. For example, roads may extend across the entire image, and traditional CNNs, constrained by their limited receptive fields, may result in fragmented predictions. Similarly, large-scale objects like forests and lakes require global contextual information to maintain classification consistency. However, CNNs may produce discontinuous regional classifications due to their reliance on local feature extraction. Directly applying CNN-based feature extraction for the global branch may fail to capture distant pixel relationships, leading to incomplete global representations. To overcome the limitations of CNN receptive fields, incorporating a self-attention mechanism before the global branch allows direct long-range pixel interactions, enhancing the model’s ability to encode global feature dependencies more effectively.

Remote sensing images typically have ultra-high resolutions, requiring the division of large images into smaller patches for processing and feature extraction. However, this local partitioning strategy may introduce information fragmentation issues, negatively impacting segmentation performance. Specifically, the lack of connectivity between image patches may lead to inconsistencies in classification, as the same object appearing in different patches may not be correctly correlated, resulting in discontinuous spatial representation of land cover features. Additionally, due to the absence of global information during patch-based processing, boundary misalignment may occur, leading to blurred or discontinuous object boundaries. After local feature extraction, a self-attention mechanism is introduced to mitigate these issues and enhance global information modeling. This mechanism enables inter-patch feature interaction, ensuring that objects of the same category maintain consistent feature representations across different patches. Furthermore, self-attention strengthens boundary feature representation, alleviating segmentation ambiguities caused by patch-wise processing and ultimately improving the clarity and accuracy of object boundaries in semantic segmentation.

It is worth noting that a fundamental concept for understanding the attention mechanism is QKV (Query et al.). In Masked cross-attention, Q and V can come from one sequence and K from another, and vice versa. This flexibility allows for adaptation to different application scenarios, including cases where queries and values of one sequence influence the attention mechanism and the keys come from another sequence.

It is worth noting that a fundamental concept for understanding the attention mechanism is QKV (Query et al.). In Masked cross-attention, Q and V can come from one sequence and K from another, and vice versa. This flexibility allows for adaptation to different application scenarios, including cases where queries and values of one sequence influence the attention mechanism and the keys come from another sequence.

\begin{figure}[H]
\begin{center}
\includegraphics[width=1\linewidth]{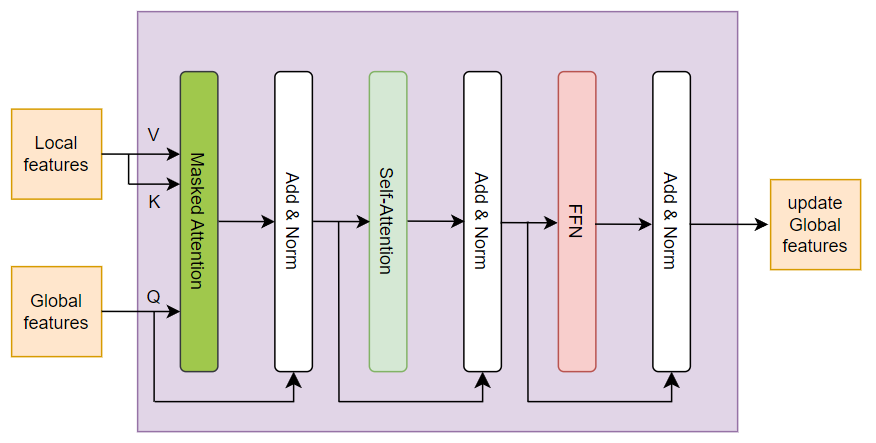}
\end{center}
	\caption{Structure of Self-Attention. Self-Attention allows the model to model relationships between different positions in the image, helping to capture long-range dependencies in computer vision tasks.
} \label{fig3}
\end{figure}
\unskip

\subsection{Global-Local Cross-Attention Fusion Module (GLCA-FM)}

The proposed Global-Local Cross-Attention Fusion Module (GLCA-FM) enhances the interaction between global semantic information and local fine-grained features, facilitating efficient cross-scale feature fusion. Specifically, this mechanism computes attention distributions between global and local features, establishing a complementary relationship across different levels. Let \(\widehat{Q_G}\) and \(\widehat{Q_L}\) represent the attention-weighted results of incorporating local information at the global level and global information at the local level, respectively. Their computations are formulated as follows:

\[
\widehat{Q_G}=\mathbf{Softmax}\left(\frac{Q_GK_L^T}{\sqrt{d_K}}\right)V_L+Q_G
\]

\[
\widehat{Q_L}=\mathbf{Softmax}\left(\frac{Q_LK_G^T}{\sqrt{d_K}}\right)V_G+Q_L
\]

Where \(Q_G\) and \(Q_L\) denote the query vectors at the global and local levels, respectively, while \(K_L\) and \(K_G\) represent the key vectors for the local and global levels, and \(V_L\) and \(V_G\) are the corresponding value vectors. The attention computation first normalizes the similarity scores between \(Q\) and \(K\) using the Softmax function, with \(\sqrt{d_K}\) serving as a scaling factor to stabilize the distribution and prevent gradient explosion or vanishing. Subsequently, the weighted summation of the value vectors yields the fused feature representations, enabling effective global-local information exchange.

This cross-attention mechanism plays a crucial role in semantic segmentation tasks. First, at the global level, \(\widehat{Q_G}\) incorporates local fine-grained details, ensuring that global features retain sensitivity to local information even at reduced resolutions, thereby enhancing global feature modeling. Second, at the local level, \(\widehat{Q_L}\) leverages guidance from global information to improve semantic consistency, reducing misclassification. Furthermore, this mechanism alleviates information fragmentation issues, particularly in patch-based processing of remote sensing imagery, where the lack of direct connections between adjacent patches may lead to inconsistent feature representations for the same object. By employing GLCA-FM, pixel relationships across patches are explicitly modeled, preserving object integrity and enhancing boundary delineation.

In summary, GLCA-FM enables deep feature fusion through bidirectional global-local interaction. Global features absorb fine-grained local details via attention weighting, while local features benefit from global semantic guidance, leading to enhanced feature stability and consistency. This bidirectional feature flow mechanism improves the integration of global and local information. It optimizes the semantic segmentation performance of ultra-high-resolution remote sensing imagery, allowing for more precise object recognition and boundary refinement in complex scenes.

\begin{figure}[H]
\begin{center}
\includegraphics[width=0.5\linewidth]{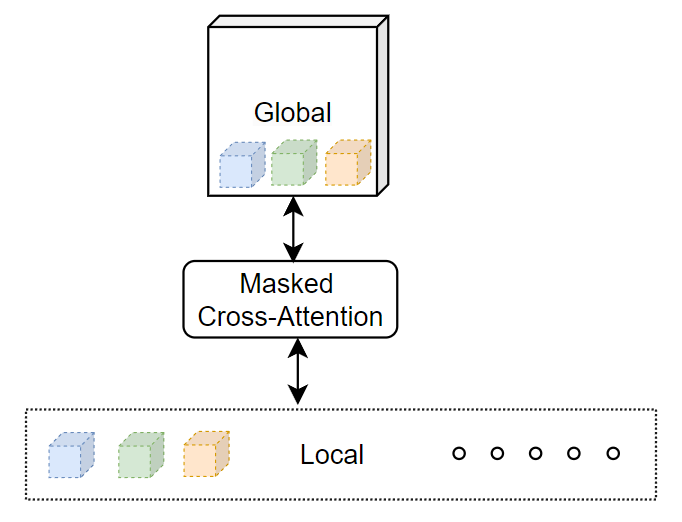}
\end{center}
\caption{Structure of Masked Cross-Attention.The Masked Cross-Attention Mechanism captures relationships between different regions in the input feature maps, enhancing multi-scale feature integration and spatial consistency. This mechanism enables the model to effectively handle complex scenes and improve segmentation accuracy, particularly in high-resolution remote sensing imagery.}
\label{fig:3}
\end{figure}

\subsection{Branch Aggregation with Regularization}
The two branches in the GLNet architecture are aggregated by an aggregation layer denoted as \( f_{\text{agg}} \), which is implemented as a 3x3 convolutional layer. This layer receives high-level feature maps \(\hat{X}_{\text{Loc}^L} \) from the \(L\) level of the local branch, as well as feature maps \(\hat{X}_{\text{Glb}^L} \) from the same level of the global branch. These feature maps are spliced along the channel and passed through the \( f_{\text{agg}} \) layer to produce the final split output \(\hat{S}_{\text{Agg}} \).

In addition to the primary segmentation loss applied to the \(\hat{S}_{\text{Agg}}\) output, two auxiliary losses are also applied to ensure that the segmentation outputs of the local branch \(\hat{S}_{\text{Loc}}\) and global branch \(\hat{S}_{\text{Glb}}\) are close to their corresponding segmentation maps (i.e., local patches and global downsampling). These auxiliary losses help stabilize the training process and ensure high segmentation accuracy.

Overall, the GLNet architecture, with its feature map sharing, deep feature map aggregation, and auxiliary loss mechanisms, has proven highly effective in achieving accurate and efficient segmentation of remote sensing images, even at ultra-high resolutions.

In practical applications, we have observed that local branches in the GLNet architecture tend to overfit some vital local details and may "override" the learning of global branches. To address this issue, we have added a weakly coupled regularization between the last layers of the two branches, which is normalized to a size of 2448 px × 500 px. Specifically, we have added a Euclidean paradigm penalty denoted as \( \lambda \|\hat{X}_{\text{Loc}^L} - \hat{X}_{\text{Glb}^L}\|_2 \) to prevent the relative change between \(\hat{X}_{\text{Loc}^L} \) and \(\hat{X}_{\text{Glb}^L} \) from being too large. Here, \(\lambda\) is a hyperparameter that is empirically fixed at 0.15 in our work.

The primary aim of this regularization is to slow down the training of the local branches so that they are more synchronized with the learning of the global branches. It helps ensure that local branches update parameters in a more stable and synchronized manner with global branches, reducing the tendency of local branches to overfit specific local entity details. Overall, this regularization has been highly effective in improving the performance and stability of the GLNet architecture for remote sensing image segmentation.

\section{EXPERIMENTS}

\subsection{ Introduction to Dataset}
The DeepGlobe dataset is an open dataset created by the Computer Science and Artificial Intelligence Laboratory at Georgia Tech. The dataset is designed to solve three critical problems related to the Earth's surface: surface building detection, road extraction, and surface water detection. These issues are fundamental components of Earth surface understanding and analysis and provide an essential foundation for any geospatial information system (GIS) or Earth observation (EO) application, such as navigation, urban planning, and disaster response.

The dataset consists of high-resolution satellite images covering six world regions: Africa, Asia, Europe, North America, South America, and Oceania. The photos are provided in RGB format and have a resolution of 2448 x 2448 pixels. The dataset also includes corresponding ground truth labels for each image, which experts in the field have manually annotated. The annotations are provided for each task, i.e., building detection, road extraction, and surface water detection, and are provided at the pixel level.

Overall, the DeepGlobe dataset has proven to be a precious resource for researchers and practitioners in remote sensing and geospatial analysis. The dataset provides a comprehensive and diverse set of images and annotations, which can be used to train and evaluate a wide range of computer vision models for various Earth surface understanding and analysis tasks.

The DeepGlobe dataset contains annotated raw remote sensing images and Earth surface labels that can be used to train computer vision (CV) and deep learning (DL) algorithms. The dataset consists of three main parts:
Surface Building Detection: This part of the dataset includes satellite imagery and building masks that allow developers to use computer vision techniques to detect buildings on the Earth's surface.

Road Extraction: This part of the dataset contains road networks and satellite imagery designed to support the differentiation of roads on the surface through computer vision techniques.

Surface water detection: This part of the dataset includes water body masks and satellite imagery that can help detect water bodies at the surface using computer vision methods.

The dataset contains 10,146 satellite images with a size of 20448×20448 pixels, divided into training/validation/testing sets of 803/171/172 images (corresponding to 70\%/15\%/15\%) each.

The ISPRS Vaihingen dataset is a publicly available standard dataset of aerial remote sensing images provided by the International Society for Photogrammetry and Remote Sensing (ISPRS). This dataset consists of six land cover categories: impervious surfaces, buildings, low vegetation, trees, cars, etc. The dataset was captured by drones over the town of Vaihingen in Germany and includes 33 true orthophoto (TOP) images composed of three bands: infrared (IR), red (R), and green (G). Additionally, corresponding DSM (Digital Surface Model) data is provided. The DSM data represents the height of trees, buildings, and other features. Both the IRRG and DSM data are used for training and inference purposes. The average size of these images is 2494 × 2064 pixels, with a spatial resolution of 9 cm. Among them, 17 scenes are used for online accuracy assessment. 

The ISPRS Potsdam dataset is a standard dataset of aerial remote sensing images publicly released by ISPRS. This dataset includes six land cover categories: impervious surfaces, buildings, low vegetation, trees, cars, and others. The dataset was captured by drones over the town of Potsdam in Germany and comprises 38 true orthophoto (TOP) images composed of four bands: infrared (IR), red (R), green (G), and blue (B). Additionally, corresponding DSM (Digital Surface Model) data is provided. The DSM data represents the height of trees, buildings, and other features. Both the IRRGB and DSM data are used for training and inference purposes. The average size of these images is 6000 × 6000 pixels, with a spatial resolution of 5 cm.

\subsection{Implementation Details}
This study used a ResNet50\cite{he2016deep} backbone network with the structure of FPN (Feature et al.). A deep feature map-sharing strategy was adopted for the feature maps of the conv2 to conv5 blocks of ResNet50 in the bottom-up phase. This strategy was also applied to the feature maps of the top-down and smoothing phases in FPN. In the final transversal connectivity stage of the FPN, feature map regularization was introduced, and the feature maps from this stage were used for the final segmentation results. The size of the downsampling and local cropping of the global image were standardized to 500 × 500 pixels to simplify the processing. A 50-pixel overlap was added between neighboring image blocks to prevent the appearance of edge effects in the convolutional layer. Focal Loss\cite{lin2017focal} was chosen as the optimization target, where the $\gamma$ value was set to 6 for the primary and two auxiliary losses. Equal weights (1.0) were assigned to primary and secondary losses. The feature map regularization factor $\lambda$ was also set to 0.15.

To evaluate the GPU memory consumption of the model, we used the command line tool "gpustat" and set the batch size to 1 while avoiding any gradient computation. It is worth noting that we used only a single GPU card during both training and inference. In our experiments, we used the PyTorch framework\cite{paszke2016enet}. During the training of the global branch, we used the Adam optimizer\cite{kinga2015method} with hyperparameters $\beta_1 = 0.9$ and $\beta_2 = 0.999$, and a learning rate of $1\times10^{-4}$. On the other hand, when training the local branch, we set the learning rate to $2\times10^{-5}$. The batch size was set to 6 for training. All experiments were conducted on a workstation with an NVIDIA 1080Ti GPU card.

\subsection{Evaluation Metrics}
One of the metrics for evaluating the performance of semantic segmentation models is the Intersection over Union (IOU), also known as the Jaccard index. IOU measures how much the pixel region the model predicts overlaps with the actual label.

The formula is $IoU = \frac{TP}{TP+FP+FN}$. True Positive (TP) refers to the number of pixels correctly predicted as the positive class by the model, while False Positive (FP) denotes the number of pixels incorrectly predicted as positive class, and False Negative (FN) represents the number of pixels erroneously predicted as the negative class by the model.

The value of IoU ranges from 0 to 1. The closer the value is to 1, the better the overlap between the region predicted by the model and the real label and the superior performance.

\begin{table}[H]
\begin{center}
\caption{THE RESULTS ON DEEPGLOBE DATASET }
\begin{tabular}{|c|cc|cl|}
\hline
\multirow{2}{*}{Model}                            & \multicolumn{2}{c|}{Patch Inference} & \multicolumn{2}{c|}{Global Inference} \\ \cline{2-5} 
 &
  \multicolumn{1}{c|}{mIoU(\%)} &
  \begin{tabular}[c]{@{}c@{}}Memory\\ (MB)\end{tabular} &
  \multicolumn{1}{c|}{mIoU(\%)} &
  \begin{tabular}[c]{@{}l@{}}Memory\\ (MB)\end{tabular} \\ \hline
UNet[16]                                              & \multicolumn{1}{c|}{37.3}   & 949    & \multicolumn{1}{c|}{38.4}   & 5507    \\ \hline
ICNet[27]                                             & \multicolumn{1}{c|}{35.5}   & 1195   & \multicolumn{1}{c|}{40.2}   & 2557    \\ \hline
PSPNet[3]                                            & \multicolumn{1}{c|}{53.3}   & 1513   & \multicolumn{1}{c|}{56.6}   & 6289    \\ \hline
SegNet[17]                                            & \multicolumn{1}{c|}{60.8}   & 1139   & \multicolumn{1}{c|}{61.2}   & 10339   \\ \hline
DeepLabv3+[4]                                        & \multicolumn{1}{c|}{63.1}   & 1279   & \multicolumn{1}{c|}{63.5}   & 3199    \\ \hline
FCN-8s[1]                                            & \multicolumn{1}{c|}{64.3}   & 1963   & \multicolumn{1}{c|}{70.1}   & 5227    \\ \hline
\multicolumn{1}{|l|}{}                            & \multicolumn{2}{c|}{mIoU(\%)}        & \multicolumn{2}{c|}{Memory(MB)}       \\ \hline
\multicolumn{1}{|l|}{GLNet:G $\Rightarrow$ L}     & \multicolumn{2}{c|}{70.9}            & \multicolumn{2}{c|}{1395}             \\ \hline
\multicolumn{1}{|l|}{GLNet:G $\Leftrightarrow$ L} & \multicolumn{2}{c|}{71.6}            & \multicolumn{2}{c|}{1865}             \\ \hline
\multicolumn{1}{|l|}{Ours}                        & \multicolumn{2}{c|}{73.4}            & \multicolumn{2}{c|}{1766}             \\ \hline
\end{tabular}
\end{center}
\end{table}

The table shows that all models achieve higher mIoU under global inference while consuming very high GPU memory. They see a drop in memory usage in block-based reasoning and a sharp drop in accuracy. Only our approach achieves the best compromise between mIoU and GPU memory usage.

\begin{figure}[H]
\centering
\centering
\includegraphics{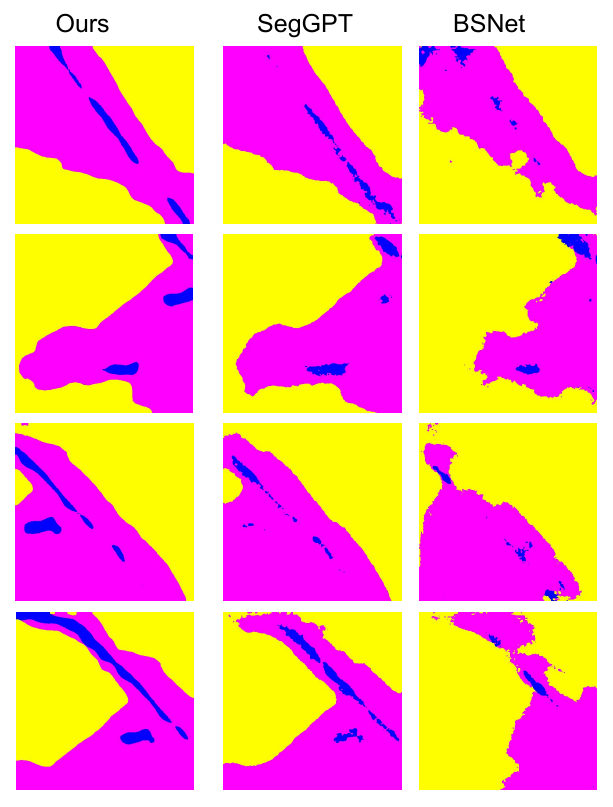}
\caption{Visualization results of our method and the best previous methods, SegGPT and BSNet. The results show that our method is significantly better than the best previous methods.}

\end{figure}

\subsection{Comparing with the existing works}

\begin{table}[H]
\begin{center}
\caption{THE RESULTS ON VAIHINGEN DATASET }
\resizebox{0.8\textwidth}{0.9in}{
\small
\begin{tabular}{cccccccc}
\hline
Method & \begin{tabular}[c]{@{}c@{}}Impervious\\ Surface\end{tabular} & Building & \begin{tabular}[c]{@{}c@{}}Low\\ Vegetation\end{tabular} & Tree & Car & OA & mIoU \\ \hline
FCN-8s[1]     & 90.0 & 93.0 & 77.7 & 86.5 & 80.4 & 85.52 & 75.5 \\ \hline
UNet[16]       & 90.5 & 93.3 & 79.6 & 87.5 & 76.4 & 85.46 & 75.5 \\ \hline
SegNet[17]     & 90.2 & 93.7 & 78.5 & 85.8 & 83.9 & 86.42 & 76.8 \\ \hline
EncNet[26]     & 91.2 & 94.1 & 79.2 & 86.9 & 83.7 & 87.02 & 77.8 \\ \hline
RefineNet[21]  & 91.1 & 94.1 & 79.8 & 87.2 & 82.3 & 86.9  & 77.1 \\ \hline
Cbam[6]       & 91.5 & 93.8 & 79.4 & 87.3 & 83.5 & 87.1  & 78.0 \\ \hline
DeepLabv3+[4] & 91.4 & 94.7 & 79.6 & 87.6 & 85.8 & 87.82 & 79.0 \\ \hline
S-RA-FCN[1]   & 90.5 & 93.8 & 79.6 & 87.5 & 82.6 & 86.8  & 77.3 \\ \hline
PSPNet[3]     & 90.6 & 94.3 & 79.0 & 87.0 & 70.7 & 84.32 & 74.1 \\ \hline
BSNet      & 92.1 & 94.4 & 83.1 & 88.3 & 86.7 & 88.92 & 80.2 \\ \hline
Mask2former & 90.4 & 93.1 & 82.7 & 79.3 & 86.0 & 86.3 & 74.9 \\ \hline
SegGPT&  91.9 & 94.7 & 83.4 & 87.0 & 86.9 & 88.7 & 80.4 \\ \hline
Ours       & 94.2 & 95.6 & 85.2 & 89.0 & 86.9 & 90.18 & 82.4     \\ \hline

\end{tabular}
}
\end{center}
\end{table}
1) Comparison of the Vaihingen dataset: Table II compares our proposed method quantitatively with state-of-the-art methods on the Vaihingen dataset. The table shows that our proposed method outperforms the other methods regarding the overall category mean intersection ratio (mIoU). FCN effectively segments large buildings but needs more attention to detail information, resulting in poor predictions of small objects such as trees and cars. S-RA-FCN aggregates long-range spatial relationships between pixels, resulting in better accuracy than FCN. UNet and SegNet improve segmentation results by using jump connections. UNet combines low-level and high-level features to restore detail, while SegNet uses a pooling index to upsample features. Additionally, the cascading residual pooling module employed by RefineNet helps to surpass the segmentation results of UNet and SegNet. This module captures background context to improve blurred segmentation at the boundaries of high spatial-resolution images. Cbam is superior to RefineNet for automobiles because it uses the height map as an additional channel to extract boundary information. In addition, Deeplabv3+ outperforms EncNet due to the contextual information captured by the null convolution, which improves boundary details. BSNet performs well in the Low Vegetation and Automotive categories. Our model achieves 94.2\%  and 95.6\% accuracy in identifying Impervious Surfaces and Buildings, respectively, significantly outperforming other methods. This reflects the superior performance of our model in handling details and complex structures. In Low Vegetation and Tree classification tasks, our model achieves 85.2\%  and 89.0\% , respectively, indicating high accuracy in recognizing different vegetation types in natural environments. However, the model performs moderately in identifying cars with 86.9\%, comparable to other methods. Our model achieves an average intersection and merger ratio of 90.18\% on the Vaihingen dataset, demonstrating excellent performance with high generalization ability and overall accuracy. The results show significant improvement in the segmentation of small and large objects, validating the effectiveness of our method.

\begin{table}[H]
\begin{center}
\caption{THE RESULTS ON POTSDAM DATASET}
\resizebox{0.8\textwidth}{1in}{
\small
\begin{tabular}{cccccccc}
\hline
Method & \begin{tabular}[c]{@{}c@{}}Impervious\\ Surface\end{tabular} & Building & \begin{tabular}[c]{@{}c@{}}Low\\ Vegetation\end{tabular} & Tree & Car & OA & mIoU \\ \hline
FCN-8s[1]     & 89.9 & 93.7 & 83.0 & 85.2 & 93.5 & 89.06 & 71.7 \\ \hline
UNet[16]       & 88.2 & 91.1 & 82.8 & 84.9 & 91.6 & 87.72 & 68.5 \\ \hline
SegNet[17]     & 87.8 & 90.7 & 81.0 & 84.7 & 89.7 & 86.78 & 66.2 \\ \hline
EncNet[26]     & 91.0 & 94.9 & 84.4 & 85.9 & 93.6 & 89.96 & 73.2 \\ \hline
RefineNet[21]  & 88.1 & 93.1 & 85.6 & 86.3 & 90.3 & 88.68 & 72.6 \\ \hline
Cbam[6]       & 88.3 & 93.2 & 84.7 & 86.0 & 92.8 & 89.0  & 72.8 \\ \hline
DeepLabv3+[4] & 91.3 & 94.8 & 84.2 & 86.6 & 93.8 & 90.14 & 73.8 \\ \hline
S-RA-FCN[1]   & 90.7 & 94.2 & 83.8 & 85.8 & 93.6 & 89.62 & 72.5 \\ \hline
PSPNet[3]     & 90.7 & 94.8 & 84.1 & 85.9 & 90.5 & 89.2  & 71.7 \\ \hline
BSNet      & 92.4 & 95.6 & 86.8 & 88.1 & 94.6 & 91.5  & 77.5 \\ \hline
Mask2Former      & 89.0 & 93.7 & 85.9 & 88.4 & 94.9 & 90.38 & 74.0 \\ \hline
SegGPT      & 93.0 & 94.3 & 85.5 & 87.2 & 93.9 & 90.24 & 78.2 \\ \hline
Ours       & 94.7 & 97.  & 87.4 & 89.0 & 95.1 & 92.64&80.4 \\ \hline

\end{tabular}
}
\end{center}
\end{table}

2) Comparison of the Potsdam dataset: To further validate our method's superiority, we conducted experiments on the Potsdam benchmark dataset. As summarized in Table III, firstly, our model performs best in identifying impervious surfaces with 94.7\%, which is a significant improvement compared to other methods. It may reflect our model's ability to deal with complex surfaces in urban landscapes with superior capture of detail and structure. Second, for the classification task of buildings, our model performs equally well, achieving an accuracy of 97.0\%. This shows the model's ability to learn and generalize efficiently for complex structures and textures, a significant advantage over other methods. In the low vegetation and tree recognition tasks, our model achieved 87.4\%  and 89.0\%  accuracy, respectively, slightly higher than other methods. It indicates that our model is good at recognizing different vegetation types. For the classification task of automobiles, our model performs 95.1\%, which is the leading performance among all methods, showing its excellent performance in vehicle detection. Ultimately, our model achieves 92.64\%  in terms of mean intersection and merger ratio (me), significantly outperforming other methods. It further demonstrates the excellent performance of our method overall for all types of feature segmentation tasks. The experimental results for the Potsdam dataset show that our model performs well in semantic segmentation tasks in complex urban environments, and its ability to discriminate between different feature types efficiently makes it a strong contender in the field. However, further work can improve the model performance by profoundly analyzing the error cases and exploring more sophisticated feature extraction strategies.

In addition to numerical evaluation, we need to further demonstrate the capability of our method through visual inspection. As shown in Fig. 4, we randomly selected a test image for semantic segmentation. Overall, the results are highly discriminative and the negative effects of the edge parts are suppressed. The visual results are consistent with the numerical evaluation. Our method effectively outlines the edge details in large-scale satellite images, which cannot be captured by other methods.

\subsection{Ablation Study}
\begin{table}[!h]
\begin{center}
\caption{ABLATION EXPERIMENT }
\begin{tabular}{|c|c|c|c|}
\hline
Method                & Mask    & Self-attention & mIoU \\ \hline
\multirow{3}{*}{Ours} & $\surd$ &                & 73.0 \\ \cline{2-4} 
                      &         & $\surd$        & 72.4 \\ \cline{2-4} 
                      & $\surd$ & $\surd$        & 73.4 \\ \hline
\end{tabular}
\end{center}
\end{table}

The experiments conducted for Mask-Attention and Self-Attention involved various settings to verify their performance. In comparison to the proposed approach, the category mean intersection ratio (mIoU) was observed to be 72.4\% when Mask-Attention was not used, 73.0\% when Self-Attention was not used and improved to 73.4\% when both Mask-Attention and Self-Attention were used. These results demonstrate the effectiveness of the proposed method in achieving higher precision semantic segmentation.

\section{CONCLUSION}
This paper introduces an efficient two-branch segmentation model for ultra-high-resolution remote sensing images. The proposed model, GLCANet, employs a Global-Local Cross-Attention (GLCA-FM) feature fusion module to effectively integrate global and local features, enhancing segmentation accuracy and detail preservation. GLCANet enables refined feature representation and improved boundary delineation by incorporating Self-Attention and Masked-Attention mechanisms. Experimental evaluations across multiple datasets demonstrate that GLCANet surpasses existing methods in segmentation performance. However, its generalization ability across diverse datasets remains to be fully validated, as the current experiments were conducted on a limited set of datasets. Furthermore, due to its cross-attention mechanism and dual-branch architecture, GLCANet is computationally more intensive, requiring excellent training time and hardware resources. The experiments were conducted using a single GPU, which may have constrained the model’s training efficiency and inference speed.
\bibliographystyle{plain} 
\bibliography{reference1}

\begin{thebibliography}{10}

\bibitem{badrinarayanan2017segnet}
Vijay Badrinarayanan, Alex Kendall, and Roberto Cipolla.
\newblock Segnet: A deep convolutional encoder-decoder architecture for image segmentation.
\newblock {\em IEEE transactions on pattern analysis and machine intelligence}, 39(12):2481--2495, 2017.

\bibitem{bazi2021vision}
Yakoub Bazi, Laila Bashmal, Mohamad M~Al Rahhal, Reham~Al Dayil, and Naif~Al Ajlan.
\newblock Vision transformers for remote sensing image classification.
\newblock {\em Remote Sensing}, 13(3):516, 2021.

\bibitem{bonafilia2020sen1floods11}
Derrick Bonafilia, Beth Tellman, Tyler Anderson, and Erica Issenberg.
\newblock Sen1floods11: A georeferenced dataset to train and test deep learning flood algorithms for sentinel-1.
\newblock In {\em Proceedings of the IEEE/CVF conference on computer vision and pattern recognition workshops}, pages 210--211, 2020.

\bibitem{chen2021crossvit}
Chun-Fu~Richard Chen, Quanfu Fan, and Rameswar Panda.
\newblock Crossvit: Cross-attention multi-scale vision transformer for image classification.
\newblock In {\em Proceedings of the IEEE/CVF international conference on computer vision}, pages 357--366, 2021.

\bibitem{chen2021remote}
Hao Chen, Zipeng Qi, and Zhenwei Shi.
\newblock Remote sensing image change detection with transformers.
\newblock {\em IEEE Transactions on Geoscience and Remote Sensing}, 60:1--14, 2021.

\bibitem{f2net}
Hengzhi Chen, Liqian Feng, Wenhua Wu, Xiaogang Zhu, Shawn Leo, and Kun Hu.
\newblock F2net: A frequency-fused network for ultra-high resolution remote sensing segmentation.
\newblock {\em arXiv preprint arXiv:2506.07847}, 2025.

\bibitem{chen2014semantic}
Liang-Chieh Chen, George Papandreou, Iasonas Kokkinos, Kevin Murphy, and Alan~L Yuille.
\newblock Semantic image segmentation with deep convolutional nets and fully connected crfs.
\newblock {\em arXiv preprint arXiv:1412.7062}, 2014.

\bibitem{chen2017deeplab}
Liang-Chieh Chen, George Papandreou, Iasonas Kokkinos, Kevin Murphy, and Alan~L Yuille.
\newblock Deeplab: Semantic image segmentation with deep convolutional nets, atrous convolution, and fully connected crfs.
\newblock {\em IEEE transactions on pattern analysis and machine intelligence}, 40(4):834--848, 2017.

\bibitem{chen2018encoder}
Liang-Chieh Chen, Yukun Zhu, George Papandreou, Florian Schroff, and Hartwig Adam.
\newblock Encoder-decoder with atrous separable convolution for semantic image segmentation.
\newblock In {\em Proceedings of the European conference on computer vision (ECCV)}, pages 801--818, 2018.

\bibitem{ding2020lanet}
Lei Ding, Hao Tang, and Lorenzo Bruzzone.
\newblock Lanet: Local attention embedding to improve the semantic segmentation of remote sensing images.
\newblock {\em IEEE Transactions on Geoscience and Remote Sensing}, 59(1):426--435, 2020.

\bibitem{ding2021adversarial}
Lei Ding, Hao Tang, Yahui Liu, Yilei Shi, Xiao~Xiang Zhu, and Lorenzo Bruzzone.
\newblock Adversarial shape learning for building extraction in vhr remote sensing images.
\newblock {\em IEEE Transactions on Image Processing}, 31:678--690, 2021.

\bibitem{ding2016road}
Lei Ding, Qimiao Yang, Jun Lu, Junfeng Xu, and Jintao Yu.
\newblock Road extraction based on direction consistency segmentation.
\newblock In {\em Pattern Recognition: 7th Chinese Conference, CCPR 2016, Chengdu, China, November 5-7, 2016, Proceedings, Part I 7}, pages 131--144. Springer, 2016.

\bibitem{binary}
Xingyu Ding, Lianlei Shan, Guiqin Zhao, Meiqi Wu, Wenzhang Zhou, and Wei Li.
\newblock The binary quantized neural network for dense prediction via specially designed upsampling and attention.
\newblock {\em arXiv preprint arXiv:2405.17776}, 2024.

\bibitem{dosovitskiy2020image}
Alexey Dosovitskiy, Lucas Beyer, Alexander Kolesnikov, Dirk Weissenborn, Xiaohua Zhai, Thomas Unterthiner, Mostafa Dehghani, Matthias Minderer, Georg Heigold, Sylvain Gelly, et~al.
\newblock An image is worth 16x16 words: Transformers for image recognition at scale.
\newblock {\em arXiv preprint arXiv:2010.11929}, 2020.

\bibitem{rs2}
Bingyun Du, Lianlei Shan, Xiaoyu Shao, Dongyou Zhang, Xinrui Wang, and Jiaxi Wu.
\newblock Transform dual-branch attention net: Efficient semantic segmentation of ultra-high-resolution remote sensing images.
\newblock {\em Remote Sensing}, 17(3):540, 2025.

\bibitem{duan2019multiscale}
Lunhao Duan and Xiangyun Hu.
\newblock Multiscale refinement network for water-body segmentation in high-resolution satellite imagery.
\newblock {\em IEEE Geoscience and Remote Sensing Letters}, 17(4):686--690, 2019.

\bibitem{he2016deep}
Kaiming He, Xiangyu Zhang, Shaoqing Ren, and Jian Sun.
\newblock Deep residual learning for image recognition.
\newblock In {\em Proceedings of the IEEE conference on computer vision and pattern recognition}, pages 770--778, 2016.

\bibitem{ldnet}
Yuyang Ji and Lianlei Shan.
\newblock Ldnet: Semantic segmentation of high-resolution images via learnable patch proposal and dynamic refinement.
\newblock In {\em 2024 IEEE International Conference on Multimedia and Expo (ICME)}, pages 1--6. IEEE, 2024.

\bibitem{kinga2015method}
D~Kinga, Jimmy~Ba Adam, et~al.
\newblock A method for stochastic optimization.
\newblock In {\em International conference on learning representations (ICLR)}, volume~5, page~6. San Diego, California, 2015.

\bibitem{liminglong}
Minglong Li, Lianlei Shan, Xiaobin Li, Yang Bai, Dengji Zhou, Weiqiang Wang, Ke~Lv, Bin Luo, and Si-Bao Chen.
\newblock Global-local attention network for semantic segmentation in aerial images.
\newblock In {\em 2020 25th International Conference on Pattern Recognition (ICPR)}, pages 5704--5711. IEEE, 2021.

\bibitem{energy}
Xiaobin Li, Lianlei Shan, Minglong Li, and Weiqiang Wang.
\newblock Energy minimum regularization in continual learning.
\newblock In {\em 2020 25th International Conference on Pattern Recognition (ICPR)}, pages 6404--6409. IEEE, 2021.

\bibitem{fusing}
Xiaobin Li, Lianlei Shan, and Weiqiang Wang.
\newblock Fusing multitask models by recursive least squares.
\newblock In {\em ICASSP 2021-2021 IEEE International Conference on Acoustics, Speech and Signal Processing (ICASSP)}, pages 3640--3644. IEEE, 2021.

\bibitem{lin2017refinenet}
Guosheng Lin, Anton Milan, Chunhua Shen, and Ian Reid.
\newblock Refinenet: Multi-path refinement networks for high-resolution semantic segmentation.
\newblock In {\em Proceedings of the IEEE conference on computer vision and pattern recognition}, pages 1925--1934, 2017.

\bibitem{lin2017feature}
Tsung-Yi Lin, Piotr Doll{\'a}r, Ross Girshick, Kaiming He, Bharath Hariharan, and Serge Belongie.
\newblock Feature pyramid networks for object detection.
\newblock In {\em Proceedings of the IEEE conference on computer vision and pattern recognition}, pages 2117--2125, 2017.

\bibitem{lin2017focal}
Tsung-Yi Lin, Priya Goyal, Ross Girshick, Kaiming He, and Piotr Doll{\'a}r.
\newblock Focal loss for dense object detection.
\newblock In {\em Proceedings of the IEEE international conference on computer vision}, pages 2980--2988, 2017.

\bibitem{liu2020connecting}
Ding Liu, Bihan Wen, Jianbo Jiao, Xianming Liu, Zhangyang Wang, and Thomas~S Huang.
\newblock Connecting image denoising and high-level vision tasks via deep learning.
\newblock {\em IEEE Transactions on Image Processing}, 29:3695--3706, 2020.

\bibitem{liu2017image}
Ding Liu, Bihan Wen, Xianming Liu, Zhangyang Wang, and Thomas~S Huang.
\newblock When image denoising meets high-level vision tasks: A deep learning approach.
\newblock {\em arXiv preprint arXiv:1706.04284}, 2017.

\bibitem{liu2021swin}
Ze~Liu, Yutong Lin, Yue Cao, Han Hu, Yixuan Wei, Zheng Zhang, Stephen Lin, and Baining Guo.
\newblock Swin transformer: Hierarchical vision transformer using shifted windows.
\newblock In {\em Proceedings of the IEEE/CVF international conference on computer vision}, pages 10012--10022, 2021.

\bibitem{llmcot}
Hailong Luo, Bin Wu, Hongyong Jia, Qingqing Zhu, and Lianlei Shan.
\newblock Llm-cot enhanced graph neural recommendation with harmonized group policy optimization.
\newblock {\em arXiv preprint arXiv:2505.12396}, 2025.

\bibitem{geogrambench}
Shixian Luo, Zezhou Zhu, Yu~Yuan, Yuncheng Yang, Lianlei Shan, and Yong Wu.
\newblock Geogrambench: Benchmarking the geometric program reasoning in modern llms.
\newblock {\em arXiv preprint arXiv:2505.17653}, 2025.

\bibitem{dlnet}
Weijun Meng, Lianlei Shan, Sugang Ma, Dan Liu, and Bin Hu.
\newblock Dlnet: A dual-level network with self-and cross-attention for high-resolution remote sensing segmentation.
\newblock {\em Remote Sensing}, 17(7):1119, 2025.

\bibitem{noh2015learning}
Hyeonwoo Noh, Seunghoon Hong, and Bohyung Han.
\newblock Learning deconvolution network for semantic segmentation.
\newblock In {\em Proceedings of the IEEE international conference on computer vision}, pages 1520--1528, 2015.

\bibitem{paszke2016enet}
Adam Paszke, Abhishek Chaurasia, Sangpil Kim, and Eugenio Culurciello.
\newblock Enet: A deep neural network architecture for real-time semantic segmentation.
\newblock {\em arXiv preprint arXiv:1606.02147}, 2016.

\bibitem{synthetic}
Ruochen Pi and Lianlei Shan.
\newblock Synthetic lung x-ray generation through cross-attention and affinity transformation.
\newblock {\em arXiv preprint arXiv:2503.07209}, 2025.

\bibitem{ronneberger2015u}
Olaf Ronneberger, Philipp Fischer, and Thomas Brox.
\newblock U-net: Convolutional networks for biomedical image segmentation.
\newblock In {\em Medical Image Computing and Computer-Assisted Intervention--MICCAI 2015: 18th International Conference, Munich, Germany, October 5-9, 2015, Proceedings, Part III 18}, pages 234--241. Springer, 2015.

\bibitem{acmmm}
Leo Shan, Wenzhang Zhou, and Grace Zhao.
\newblock Incremental few shot semantic segmentation via class-agnostic mask proposal and language-driven classifier.
\newblock In {\em Proceedings of the 31st ACM International Conference on Multimedia}, pages 8561--8570, 2023.

\bibitem{uhrsnet}
Lianlei Shan, Minglong Li, Xiaobin Li, Yang Bai, Ke~Lv, Bin Luo, Si-Bao Chen, and Weiqiang Wang.
\newblock Uhrsnet: A semantic segmentation network specifically for ultra-high-resolution images.
\newblock In {\em 2020 25th International Conference on Pattern Recognition (ICPR)}, pages 1460--1466. IEEE, 2021.

\bibitem{decouple}
Lianlei Shan, Xiaobin Li, and Weiqiang Wang.
\newblock Decouple the high-frequency and low-frequency information of images for semantic segmentation.
\newblock In {\em ICASSP 2021-2021 IEEE International Conference on Acoustics, Speech and Signal Processing (ICASSP)}, pages 1805--1809. IEEE, 2021.

\bibitem{cognitive}
Lianlei Shan, Shixian Luo, Zezhou Zhu, Yu~Yuan, and Yong Wu.
\newblock Cognitive memory in large language models.
\newblock {\em arXiv preprint arXiv:2504.02441}, 2025.

\bibitem{densenet}
Lianlei Shan and Weiqiang Wang.
\newblock Densenet-based land cover classification network with deep fusion.
\newblock {\em IEEE Geoscience and Remote Sensing Letters}, 19:1--5, 2021.

\bibitem{mbnet}
Lianlei Shan and Weiqiang Wang.
\newblock Mbnet: A multi-resolution branch network for semantic segmentation of ultra-high resolution images.
\newblock In {\em ICASSP 2022-2022 IEEE International Conference on Acoustics, Speech and Signal Processing (ICASSP)}, pages 2589--2593. IEEE, 2022.

\bibitem{tgrs1}
Lianlei Shan, Weiqiang Wang, Ke~Lv, and Bin Luo.
\newblock Class-incremental learning for semantic segmentation in aerial imagery via distillation in all aspects.
\newblock {\em IEEE Transactions on Geoscience and Remote Sensing}, 60:1--12, 2021.

\bibitem{tgrs2}
Lianlei Shan, Weiqiang Wang, Ke~Lv, and Bin Luo.
\newblock Class-incremental semantic segmentation of aerial images via pixel-level feature generation and task-wise distillation.
\newblock {\em IEEE Transactions on Geoscience and Remote Sensing}, 60:1--17, 2022.

\bibitem{boosting}
Lianlei Shan, Weiqiang Wang, Ke~Lv, and Bin Luo.
\newblock Boosting semantic segmentation of aerial images via decoupled and multilevel compaction and dispersion.
\newblock {\em IEEE Transactions on Geoscience and Remote Sensing}, 61:1--16, 2023.

\bibitem{edge}
Lianlei Shan, Weiqiang Wang, Ke~Lv, and Bin Luo.
\newblock Edge-guided and class-balanced active learning for semantic segmentation of aerial images.
\newblock {\em arXiv preprint arXiv:2405.18078}, 2024.

\bibitem{data}
Lianlei Shan, Guiqin Zhao, Jun Xie, Peirui Cheng, Xiaobin Li, and Zhepeng Wang.
\newblock A data-related patch proposal for semantic segmentation of aerial images.
\newblock {\em IEEE Geoscience and Remote Sensing Letters}, 20:1--5, 2023.

\bibitem{lifelong}
Lianlei Shan, Wenzhang Zhou, Wei Li, and Xingyu Ding.
\newblock Lifelong learning and selective forgetting via contrastive strategy.
\newblock {\em arXiv preprint arXiv:2405.18663}, 2024.

\bibitem{organizing}
Lianlei Shan, Wenzhang Zhou, Wei Li, and Xingyu Ding.
\newblock Organizing background to explore latent classes for incremental few-shot semantic segmentation.
\newblock {\em arXiv preprint arXiv:2405.19568}, 2024.

\bibitem{gmm}
Chengsong Sun, Weiping Li, Xiang Li, Yuankun Liu, and Lianlei Shan.
\newblock Gmm-based comprehensive feature extraction and relative distance preservation for few-shot cross-modal retrieval.
\newblock {\em arXiv preprint arXiv:2505.13306}, 2025.

\bibitem{wang2021max}
Huiyu Wang, Yukun Zhu, Hartwig Adam, Alan Yuille, and Liang-Chieh Chen.
\newblock Max-deeplab: End-to-end panoptic segmentation with mask transformers.
\newblock In {\em Proceedings of the IEEE/CVF conference on computer vision and pattern recognition}, pages 5463--5474, 2021.

\bibitem{wang2020deep}
Jingdong Wang, Ke~Sun, Tianheng Cheng, Borui Jiang, Chaorui Deng, Yang Zhao, Dong Liu, Yadong Mu, Mingkui Tan, Xinggang Wang, et~al.
\newblock Deep high-resolution representation learning for visual recognition.
\newblock {\em IEEE transactions on pattern analysis and machine intelligence}, 43(10):3349--3364, 2020.

\bibitem{wang2018detect}
Tiantian Wang, Lihe Zhang, Shuo Wang, Huchuan Lu, Gang Yang, Xiang Ruan, and Ali Borji.
\newblock Detect globally, refine locally: A novel approach to saliency detection.
\newblock In {\em Proceedings of the IEEE conference on computer vision and pattern recognition}, pages 3127--3135, 2018.

\bibitem{wang2018non}
Xiaolong Wang, Ross Girshick, Abhinav Gupta, and Kaiming He.
\newblock Non-local neural networks.
\newblock In {\em Proceedings of the IEEE conference on computer vision and pattern recognition}, pages 7794--7803, 2018.

\bibitem{zhaoyuzhong}
Weijia Wu, Yuzhong Zhao, Zhuang Li, Lianlei Shan, Hong Zhou, and Mike~Zheng Shou.
\newblock Continual learning for image segmentation with dynamic query.
\newblock {\em IEEE Transactions on Circuits and Systems for Video Technology}, 34(6):4874--4886, 2023.

\bibitem{xie2021segformer}
Enze Xie, Wenhai Wang, Zhiding Yu, Anima Anandkumar, Jose~M Alvarez, and Ping Luo.
\newblock Segformer: Simple and efficient design for semantic segmentation with transformers.
\newblock {\em Advances in neural information processing systems}, 34:12077--12090, 2021.

\bibitem{geolocsft}
Qiang Yi and Lianlei Shan.
\newblock Geolocsft: Efficient visual geolocation via supervised fine-tuning of multimodal foundation models.
\newblock {\em arXiv preprint arXiv:2506.01277}, 2025.

\bibitem{flexdataset}
Ellen Yi-Ge and Leo Shawn.
\newblock Flexdataset: Crafting annotated dataset generation for diverse applications.
\newblock In {\em Proceedings of the AAAI Conference on Artificial Intelligence}, volume~39, pages 9481--9489, 2025.

\bibitem{yu2018bisenet}
Changqian Yu, Jingbo Wang, Chao Peng, Changxin Gao, Gang Yu, and Nong Sang.
\newblock Bisenet: Bilateral segmentation network for real-time semantic segmentation.
\newblock In {\em Proceedings of the European conference on computer vision (ECCV)}, pages 325--341, 2018.

\bibitem{yu2015multi}
Fisher Yu and Vladlen Koltun.
\newblock Multi-scale context aggregation by dilated convolutions.
\newblock {\em arXiv preprint arXiv:1511.07122}, 2015.

\bibitem{zhang2018context}
Hang Zhang, Kristin Dana, Jianping Shi, Zhongyue Zhang, Xiaogang Wang, Ambrish Tyagi, and Amit Agrawal.
\newblock Context encoding for semantic segmentation.
\newblock In {\em Proceedings of the IEEE conference on Computer Vision and Pattern Recognition}, pages 7151--7160, 2018.

\bibitem{zhang2018road}
Zhengxin Zhang, Qingjie Liu, and Yunhong Wang.
\newblock Road extraction by deep residual u-net.
\newblock {\em IEEE Geoscience and Remote Sensing Letters}, 15(5):749--753, 2018.

\bibitem{zhaoguiqin}
Guiqin Zhao, Lianlei Shan, and Weiqiang Wang.
\newblock End-to-end remote sensing change detection of unregistered bi-temporal images for natural disasters.
\newblock In {\em International Conference on Artificial Neural Networks}, pages 259--270. Springer, 2023.

\bibitem{zhao2018icnet}
Hengshuang Zhao, Xiaojuan Qi, Xiaoyong Shen, Jianping Shi, and Jiaya Jia.
\newblock Icnet for real-time semantic segmentation on high-resolution images.
\newblock In {\em Proceedings of the European conference on computer vision (ECCV)}, pages 405--420, 2018.

\bibitem{zhao2024cooperative}
Kai Zhao and Wei Xiong.
\newblock Cooperative connection transformer for remote sensing image captioning.
\newblock {\em IEEE Transactions on Geoscience and Remote Sensing}, 62:1--14, 2024.

\bibitem{dynrsl}
Xirui Zhou, Lianlei Shan, and Xiaolin Gui.
\newblock Dynrsl-vlm: Enhancing autonomous driving perception with dynamic resolution vision-language models.
\newblock {\em arXiv preprint arXiv:2503.11265}, 2025.

\end{thebibliography}

\end{document}